# Correlating and Cross-linking Knowledge Threads in Informledge System for Creating New Knowledge

T.R. Gopalakrishnan Nair[1,2] and Meenakshi Malhotra[3,4]

[1]*Saudi Aramco Endowed Chair, Technology and Information Management, PMU. KSA*
[2]*Vice President Aadvanced AI and Bio Computing, RIIC, D. S. Institutions,Bangalore, India*
[3] *Advanced AI and Bio Computing, RIIC, D. S. Institutions,* Bangalore, India
[4]*(Lecturer): Dept. of Computer Science Eng, The Oxford College of Engineering,* Bangalore, India
*trgnair@ieee.org, uppal_meenakshi@yahoo.co.in*

Keywords: Informledge System (ILS): Knowledge Network Node (KNN): Multi-lateral links: weighted graphs: Tensor graph product: Concept: Concept State Diagram (CSD)

Abstract: There has been a considerable advance in computing, to mimic the way in which the brain tries to comprehend and structure the information to retrieve meaningful knowledge. It is identified that neuronal entities hold whole of the knowledge that the species makes use of. We intended to develop a modified knowledge based system, termed as Informledge System (ILS) with autonomous nodes and intelligent links that integrate and structure the pieces of knowledge. We conceive that every piece of knowledge is a cluster of cross-linked and correlated structure. In this paper, we put forward the theory of the nodes depicting concepts, referred as Entity Concept State which in turn is dealt with Concept State Diagrams (CSD). This theory is based on an abstract framework provided by the concepts. The framework represents the ILS as the weighted graph where the weights attached with the linked nodes help in knowledge retrieval by providing the direction of connectivity of autonomous nodes present in knowledge thread traversal. Here for the first time in the process of developing Informledge, we apply tenor computation for creating intelligent combinatorial knowledge with cross mutation to create fresh knowledge which looks to be the fundamentals of a typical thought process.

## 1 INTRODUCTION

With the advancement in the field of artificial intelligence and cognitive science, development of knowledge based systems has taken a greater leap. Artificial Intelligence as stated is the study of the computations that make it possible to perceive reasons and act (Winston, Patrick Henry, 1992). Cognitive Science as referred by Thagard, Paul (2005) is the field of study of mind and its computational intelligence. Researchers from both the fields have been striving to develop a knowledge based system that could simulate the mode of existence or either of human knowledge and human information processing mechanism or both.

Knowledge management field developed so far involve information storage, processing and retrieval of information stored in databases in the form of fixed sentences and words. At present there is no meaningful system at hand that can store elements of concepts from different domains coherently and can retrieve correlated knowledge. To overcome this, we had suggested Informledge System (ILS) in our work published earlier (T.R. Gopalakrishnan Nair, Meenakshi Malhotra, 2011 and 2010). In ILS, knowledge units belonging to same or different knowledge domains are linked together to form a knowledge network using the autonomous nodes and intelligent links. These links play an important role in connecting correlated concepts stored at the knowledge units. The knowledge unit in ILS is termed as Knowledge Network Node (KNN) as defined by T.R. Gopalakrishnan Nair, Meenakshi Malhotra (2010).In this paper we put forward the theory of concepts represented by KNNs and the Concept State Diagram (CSD) depicted by the knowledge thread. This knowledge thread, formed during knowledge embedding and retrieval, is the result of intelligent linking of KNNs, wherein every link formed is composed of multiple strands each of which hold a property (T.R. Gopalakrishnan Nair, Meenakshi Malhotra, 2011).

This paper is organized as follows. Section II discusses Research Background. It is then followed by a brief introduction to the understanding of the theory of concept in section III. Section IV gives the Structural Organization of Concepts and CSD and in section V we discuss the formation of knowledge threads in ILS and section VI focuses on details about Tensor algebra applied to the CSD. Finally, we conclude the work in section VII.

## 2 RESEARCH BACKGROUND

As believed, all the information processing for human cognition is held at the neurons. A neuron receive information, process it and forward the control to another neuron for the further information processing, in the interconnected mesh of neurons (David Sánchez, 2010). In almost similar process ILS, the KNN receive the inputs and have the inbuilt ability to infer and reason the linkages to the other nodes (T.R. Gopalakrishnan Nair, Meenakshi Malhotra, 2011).

In addition to this, knowledge-based neurocomputing has gained importance in last two decades. It is stated that Knowledge-based neurocomputing (KBN) concerns with methods to address the explicit representation and processing of knowledge where a neurocomputing system is involved (Ian Cloete and Jacek M. Zurada, 2000).

The key element involved in knowledge processing and retrieval is the knowledge and its representation. Knowledge representation has been recognized as an imperative field of artificial intelligence which involves information embedding and processing for computation in cognitive models. Knowledge has been represented using network, graphs, and finite automata and using concept maps. According to Christopher Brewster (2004) many knowledge based representations involve use of Ontology. Ontology finds its origin from the field of philosophy whereas its implication in the field of computer science is stated as "ontology is formal, explicit specification of a shared conceptualization" (Thomas R. Gruber, 1993). Also, there have been efforts to define the set of attributes for the concepts involved during ontology development (Priss, U., 2006.).

Different strategies have been adopted to represent concepts. Concept maps are used to represent and convey knowledge. It is a diagram that connects pieces with of information entities that are linked by labelled straight lines without any processing power. ''Mind Maps'' are such a type of meaning diagram as shown by Beth Crandall et al. (2006) and Farrand, P et al. (2002).The connecting lines in Mind Map are not labelled and they represent just the connection between ideas (Open Directory - Reference: Knowledge Management, 2009). Also, the diagrams that are referred to as ''Cognitive Maps'' are large web like diagrams which involve representation of sentences and short paragraphs as ideas resulting in hundreds of joins and the same has been shown with example by (Robert M. Kitchin, 1994). However these maps are just the fixed representation of joints representing words whereas the knowledge representation does not involve the language alone. Similarly, the knowledge formation in human brain includes concept formation and its representation in different regions as stated in the field of neuroscience (Eric R. Kandel et al., 2000).While the mind maps connect the ideas they are not capable of processing the knowledge nodes, but in the case of ILS, the system is intelligent by virtue of its capability to formulate and process the concepts which is kept apart from how the knowledge is represented (Nair, T.R.G., Malhotra, Meenakshi, 2011).

Conceptual graph is a graphical representation of knowledge depiction and reasoning. The translation to and from the spoken language, used for understanding, into some computer understandable representation can be done by means of these graphs (Sowa, John F., 1984). As stated by Michel Chein , Marie-Laure Mugnier (2008), conceptual graph involve mainly relations between concepts as "is a" and "has property". This restricts the way in which knowledge can be linked, whereas ILS provides the flexibility in linking properties of the KNNs through the use of its multi-stranded links.

## 3 CONCEPT THEORY

Knowledge is nothing but a collection of linked concepts. However many attempted to create knowledge bases, connect words and sentences rather than concepts (Rajendra Akerkar, Priti Sajja, 2010). In his book, Gregory L. Murphy (2002) referred that many properties of concepts are found in word-meaning and use, suggesting that meanings are psychologically represented through the conceptual system. Most of this type of approach ended connecting words through lines where as the meaning of the knowledge structure is created in the human mind. But actual breakthrough is required in incorporating a meaning creation processing capability in the nodes and in the links by adding intelligence in both. In ILS, the information is stored at the intelligent knowledge network nodes (KNN) which is an integral part of this knowledge network. A concept is represented partially or completely by a KNN. The link between these autonomous nodes represents the affinity between the concepts. For the loosely related concepts the thickness of the link is less whereas for tightly coupled concepts the thickness is more. As more and more information is added into the network, there is cross-linking of concepts and new knowledge threads can be retrieved through the interlinked concepts correlation and cross-linking. In the first case of existing knowledge bases, they were static representation of the thinking process and where as ILS is capable of furnishing active computation process.

## 4  CONCEPT FRAMEWORK

Concept State Diagram (CSD) is the diagram that depicts the states of the concepts that are connected by strengthening of the links. CSD is basically a knowledge thread where state is associated with the concepts as shown in Figure 1.

The knowledge that can be retrieved after incorporating knowledge representation in to the retrieved knowledge threads is anomalous to the sentence in language. The knowledge thread formed by connecting concepts have two basic knowledge units namely Apex Knowledge Unit (AKU) and Subsidiary Knowledge Unit (SKU).

CSD, which is a form of a knowledge thread comprise of these knowledge units defined as follows:

### 4.1  Apex Knowledge Unit (AKU)

This knowledge unit in the CSD is the key concept which implies that the knowledge thread formed by CSD has this knowledge unit (KNN) representing the concept about which something is stated in the knowledge thread. E.g. considering the knowledge thread "Continent is the largest and continuous landmass on earth" can be represented in CSD as shown in Figure 1.

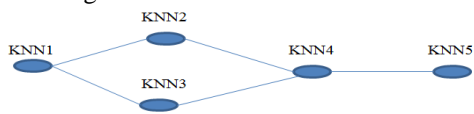

Figure 1: Concept State Diagram for a Knowledge Thread.

Where the KNN's shown in figure, 1 holds the concepts shown in table 1 below

Table 1: Concepts and KNNs shown in figure 1.

| KNN | KNN1 | KNN2 | KNN3 | KNN4 | KNN5 |
|---|---|---|---|---|---|
| Concept | Continent | Largest | Continuous | Landmass | Earth |

The AKUs here can be continent (KNN1), landmass (KNN4) or earth (KNN5). The centre concept for this CSD can be taken as any of the three. The weight attached with any of the AKU would be more than the weight attached with the nodes that are far from the AKU. The weight attached with interlinked concepts decreases as we traverse from AKU to SKU.

### 4.2  Subsidiary Knowledge Unit (SKU)

This knowledge unit in the CSD is an auxiliary concept which implies that the knowledge thread formed by CSD has these knowledge units (KNNs) in the form of supplementary concept. E.g. considering the CSD as shown in Figure 1, largest (KNN2) and continuous (KNN3) are the SKUs. These SKUs supplement the AKUs, in common language representation SKUs represent the predicates whereas AKU's represent subjects.

Hence the AKU and SKU provides a much wider representation for knowledge in ILS analogous to subject and predicates used in human language representation.

## 5  CONCEPT FORMATION

ILS is an organized knowledge network, in which connectivity and search is done through an ordered set of links. Methods of understanding is in links which includes classifying, correlating and extrapolating the information. Properties of links leads to processing where by several logically interlinked concepts can be retrieved through it. The interlinked links form CSDs, which facilitate to form the cluster to knowledge threads during retrieval. Before we put forward how this knowledge cluster is formed, we brief about the CSDs.

The necessary and sufficient conditions for CSD formation are stated as follows:

## 5.1 Every CSD that is formed should consist of at least one AKU and several SKUs.

This can be proved by contradiction, by including the following two cases:

- Let us presume CSD doesn't have an AKU
  Assume that there is a CSD that doesn't have an AKU. Let's consider such a CSD shown in figure 2, for this CSD there is no AKU specified during embedding of this knowledge. So this implies that all KNNs linked would be SKUs only.
  As stated earlier AKU and SKU are like subject and predicates, so AKU could be similar to what the knowledge thread is about. Absence of AKU would imply that some concepts represented by SKUs are linked but doesn't provide any information about anything. Such a CSD is termed as irrelevant and meaningless.

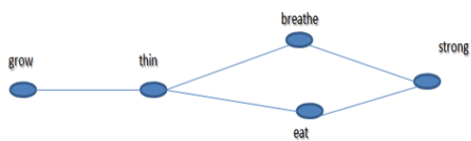

Figure 2: Incorrect CSD without AKU.

As shown in figure 2 the concepts grow, thin, breathe, eat and strong, are just linked together but no information can be retrieved from their linking. Also, in ILS links properties provide weightage which tend to present weak threads linking these concepts.

- Let us presume CSD have multiple AKU's and no SKU
  On the other hand if no SKUs are there then it's similar to a sentence not having any predicates. This would mean some subjects are linked, leading to no bigger concept. Main objective behind linking concepts is to form a bigger concept, whereas linking only AKU would not provide any meaningful knowledge. Same is depicted in the figure 3 where animal, plant, stick, door and hands are linked but this CSD does not provide any meaningful knowledge.

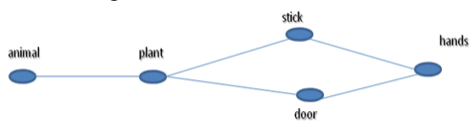

Figure 3: Incorrect CSD with multiple AKU and no SKU.

These AKUs when put together doesn't lead to formation of any bigger concept or a valid knowledge thread. These two points can be considered similar to some invalid human knowledge formation wherein one speaks about the properties of an entity and not the entity itself or someone connects some entities with weak links. Thus, every CSD should consist of least one AKU and one or more SKUs.

## 5.2 CSDs interact to form a cluster of correct and incorrect CSDs.

During knowledge retrieval, the link processor at KNN helps in retrieving the knowledge threads. The intelligence of the system lies in interaction of these retrieved knowledge threads. CSDs interaction is basically interaction between two knowledge threads that can connect and form multiple CSDs in the form of knowledge cluster depending on the state of individual knowledge units. The CSDs that are formed by CSD interaction may be correct and incorrect.

## 5.3 The CSD formed after learning has got only one form.

During knowledge retrieval, CSDs interact to form a cluster of valid and invalid CSDs. In addition to having invalid CSDs, the cluster comprises of repeated connected concepts based on the weights of the concepts involved. The learning process involves removal of invalid and the redundant CSDs. So after the learning process, which is an integral part of retrieval, is complete only one form of the CSD would be there in the knowledge network.

## 5.4 The final CSD must be conceptually correct

It is stated for the proof of Statement 3 that the invalid CSDs are removed during learning process. Hence the final CSD in the ILS are conceptually correct.

To retrieve knowledge intelligently from this network, concepts need to be combined during retrieval. These concepts interact to form combination of ideas, which is possible due to interaction between KNNs. During this interaction various links are formed which comprise to formation of both correct and incorrect CSDs. The incorrect CSDs are discarded based on the rationality involved in learning. CSDs are formed during knowledge embedding and retrieval.

The splinters retrieved during the interaction of CSDs have to pass through learning process. The process includes sieving out of the CSDs that are secondary and redundant invalid CSDs. Removal of the same from the system, lead to consistent knowledge in ILS knowledge network.

# 6 CSD AND TENSOR ALGEBRA

CSD's once formed interact during retrieval to form the cluster. This involves tensor algebra from the graph theory. In view of tensor algebra, as defined by Sandi Klavžar and Simone Severini (2010)

The tensor product, $K = G \otimes H$, of graphs $G = (V(G), E(G))$ and $H = (V(H), E(H))$ is the graph with vertex set $V(K) = V(G) \times V(H)$ and $\{(g, h), (g', h')\} \in E(K)$ if and only if $\{g, g'\} \in E(G)$ and $\{h, h'\} \in E(H)$.( p. 3)

Now as a CSD is just a graph of nodes connected together we can apply tensor product to the same. This implies that if two CSDs interact they can actually form a tensor product and the resultant graph is a cluster of CSDs composed of relevant and irrelevant CSDs. This cluster has a group of concepts which are collection of AKUs and SKUs.

Let's consider two knowledge threads kt1 and kt2 as shown in the figure 4 and figure5.

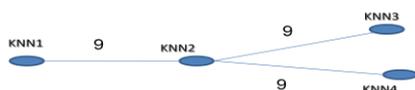

Figure 4: Knowledge Thread kt1

For the knowledge thread, kt1 shown in figure 4 represent the following knowledge and the concepts.

Knowledge thread, kt1 is "World has living and non-living things."

Table 2: Concepts represented by KNNs shown in figure 4

| KNN | Knn1 | Knn2 | Knn3 | Knn4 |
|---|---|---|---|---|
| Concept | World | Thing | Living | Non-living |

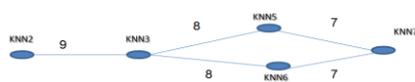

Figure 5: Knowledge Thread, kt2

The knowledge thread, Kt2 is "Living things grow and breathe like animals".

For the knowledge thread, kt2 shown in figure 5 represent the following knowledge and the concepts.

Table 3: Concepts represented by KNNs shown in figure 5

| KNN | KNN2 | KNN3 | KNN5 | KNN6 | KNN7 |
|---|---|---|---|---|---|
| Concept | Thing | Living | Grow | Breathe | Animals |

Here kt1 is similar to G and kt2 is similar to H and the vertices set for both are as follows:
V (kt1) = {KNN1, KNN2, KNN3, KNN4}

V (kt2) = {KNN2, KNN3, KNN5, KNN6, KNN7}

V (kt1) $\otimes$ V (kt2) = 4*5=20 vertices, which comprise of the node space after the 2 knowledge threads interact.

The cluster of CSDs formed after applying tensor algebra for product of connected graphs is applied to kt1 and kt2 shown in figure 4 and figure 5 respectively, is given in figure 6.

The edges connecting nodes of clusters combine to form the link space as shown in figure 6.

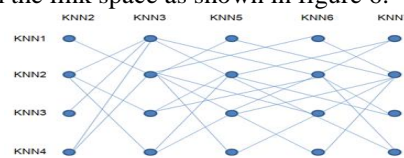

Figure 6: CSD Cluster formed after interaction of two knowledge threads, kt1 and kt2

From the definition of tensor product of graphs, the edge with vertices (KNN1, KNN2 (kt2)) and (KNN2 (knn1), KNN3) in the cluster exist if and only if (KNN1, KNN2 (kt1) is an edge (link) in kt1 and (KNN2 (kt2), KNN3) is an edge (link) in kt2. In this cluster many CSD's are correct or valid and many are incorrect or invalid. E.g.: KNN2—KNN3---KNN6 are connected which means concepts living, thing, breathe are linked to make a bigger concept which is valid. Whereas KNN4 --- KNN6 are connected which mean concept non-living and breathe are linked which is invalid.

As we cannot multiply word here and there and absurd things cannot connect. Whereas in comparison to this we can multiply two concepts or linked concepts are multiplied using tensor algebra to form a bigger concept. So it's only the concepts that are involved in the interaction which result into formation of a bigger valid concepts represented by CSDs.

## 7 CONCLUSION

In this paper, we presented the theory of KNNs which are capable of actively representing concepts and the knowledge thread as Concept State Diagram (CSD). The intelligence of ILS is depicted in formulating correlated and cross-linked cluster of CSDs through tensorial interaction of the CSDs. This correlation and cross-linking is established by the tensor algebraically manipulations. Thus ILS is capable to classifying and extrapolating the two knowledge threads retrieved and can derive all the cross-linked knowledge threads out of the interaction. On forced interaction between two tensorial knowledge threads, it can form new clusters that consist of relevant and irrelevant CSDs, from which the irrelevant CSDs can be sieved out during learning process or so to say a thinking process. The tensor algebra has been applied for CSDs manipulations on the CSDs retrieved from a single domain and this arrangement of manipulation of CSDs for multiple domain need to be evaluated in the next phase.

Acknowledgment: Authors express their sincere thanks to the IP managers, for the patenting process on the sequence of theories related to this.